\documentclass{article}

\usepackage{arxiv}

\usepackage[utf8]{inputenc} % allow utf-8 input
\usepackage[T1]{fontenc}    % use 8-bit T1 fonts
\usepackage{hyperref}       % hyperlinks
\usepackage{url}            % simple URL typesetting
\usepackage{booktabs}       % professional-quality tables
\usepackage{amsmath,amssymb,amsfonts}
\usepackage{mathtools}
\usepackage{algorithmic}
\usepackage[ruled,vlined]{algorithm2e}
\usepackage{textcomp}
\usepackage{comment}
\usepackage{booktabs}
\usepackage{caption}
\usepackage{subcaption}
\usepackage{nicefrac}       % compact symbols for 1/2, etc.
\usepackage{microtype}      % microtypography
\usepackage{lipsum}
\usepackage{graphicx}
\usepackage{cite}
\usepackage{multirow}
\usepackage{makecell}

\usepackage{color}

\title{AttenScribble: Attentive Similarity Learning for Scribble-Supervised Medical Image Segmentation}

\author{
 Mu Tian \\
  School of Biomedical Engineering\\
  Medical School\\
  Shenzhen University\\
  Shenzhen, China 518000 \\
  \texttt{mtian@szu.edu.cn} \\
   \And
Qinzhu Yang \\
  School of Biomedical Engineering\\
Medical School\\
Shenzhen University\\
Shenzhen, China 518000 \\
\texttt{yangqinzhu2020@email.szu.edu.cn} \\
  \And
Yi Gao \\
  School of Biomedical Engineering\\
  Medical School\\
  Shenzhen University\\
  Shenzhen, China 518000 \\
  \texttt{gaoyi@szu.edu.cn} \\
}

\begin{document}
\maketitle
\begin{abstract}
The success of deep networks in medical image segmentation relies heavily on massive labeled training data. However, acquiring dense annotations is a time-consuming process. Weakly-supervised methods normally employ less expensive forms of supervision, among which scribbles started to gain popularity lately thanks to its flexibility. However, due to lack of shape and boundary information, it is extremely challenging to train a deep network on scribbles that generalizes on unlabeled pixels. In this paper, we present a straightforward yet effective scribble supervised learning framework. Inspired by recent advances of transformer based segmentation, we create a pluggable spatial self-attention module which could be attached on top of any internal feature layers of arbitrary fully convolutional network (FCN) backbone. The module infuses global interaction while keeping the efficiency of convolutions. Descended from this module, we construct a similarity metric based on normalized and symmetrized attention. This attentive similarity leads to a novel regularization loss that imposes consistency between segmentation prediction and visual affinity. This attentive similarity loss optimizes the alignment of FCN encoders, attention mapping and model prediction. Ultimately, the proposed FCN+Attention architecture can be trained end-to-end guided by a combination of three learning objectives: partial segmentation loss, a customized masked conditional random fields and the proposed attentive similarity loss. Extensive experiments on public datasets (ACDC and CHAOS) showed that our framework not just out-performs existing state-of-the-art, but also delivers close performance to fully-supervised benchmark. Code will be available upon publication. 
\end{abstract}

% keywords can be removed
\section{Introduction}\label{sec:intro}
% related work is included into introduction
Deep networks have demonstrated remarkable advancements in medical image segmentation \cite{tajbakhsh2020embracing} . Nonetheless, their capacity to generalize well on unseen examples relies heavily on the quantity of manual annotated training data. Obtaining sufficient labeled data can be challenging, especially for medical image segmentation, where acquiring dense annotations requires extensive expertise. 

To mitigate this burden, weakly-supervised methods have been proposed that employ less expensive forms of supervision \cite{zhou2019prior,khoreva2017simple,can2018learning,souly2017semi}, such as scribbles \cite{luo2022scribble,zhang2022shapepu,yang2022non}, boxes \cite{dai2015boxsup}, points \cite{chen2021seminar}, or image-level labels \cite{chen2022c}. Scribble supervision becomes increasingly popular lately since it allows more flexible and natural annotations while providing reasonable initial target locations. However, due to insufficient shape and boundary information, it is extremely challenging to train a deep network on scribbles that accurately generates predictions on unlabeled pixels. 

One common remedy would be to leverage pseudo labels \cite{lee2013pseudo}, that can be generated via graph based segmentation such as random walker \cite{grady2006random} to propagate scribble annotations to unlabeled pixels. Lately, more approaches tried to exploit model output to produce reliable pseudo supervision signals. For example, Scribble2Label \cite{lee2020scribble2label} proposed to consolidate model predictions along different training iterations; CycleMix \cite{zhang2022cyclemix} adopted a mixed up strategy with consistency regularization to improve data quality. A more recent work \cite{luo2022scribble} introduced an elegantly designed dual-branch network with mixed-up strategies that achieved new state-of-the-art. Fusing predictions from separate branches introduced diversified information and thus greatly increased robustness. Aside from seeking full pseudo masks, another recent work \cite{chen2022scribble2d5} proposed to also use pre-trained edge detector to aid boundary localization. 

While pseudo label may provide richer supervision at unlabeled pixels, the impact from incorrectly generated labels can be reinforced during training \cite{obukhov2019gated}. An alternative way is to learn directly towards scribbles using a partial cross-entropy (pCE) loss \cite{lin2016scribblesup} with aided by energy minimization. For instance, the work in \cite{chen2017deeplab} applied conditional random fields (CRF) as a post-processing step to refine model predictions. More efforts have been devoted to integrating the energy term directly into end-to-end training. ScribbleSup \cite{lin2016scribblesup} and several other works \cite{tang2018normalized,tang2018regularized,veksler2020regularized} had laid the foundations for using CRF and normalized cut as an integrated learning objective. Following this path, some recent works \cite{kim2019mumford,jurdi2021surprisingly,obukhov2019gated} proposed to unify optimization steps as gradient descent in single-stage training. In particular, the work in \cite{obukhov2019gated} showed that using a straightforward CRF regularization could reach the apex in weakly supervised segmentation.

One key reason for the successes of CRF \cite{obukhov2019gated} is that it can preserve target boundaries on top of model estimates with shallow vision features, usually image intensity and spatial location. However, hand-crafted configurations of gaussian kernels, similarity metric and kernel sizes could heavily rely on expertise and dataset characteristics. In addition, though CRF, in its formulation, should possess global information, people normally constrain kernel computation within a small neighborhood for efficiency. The current CRF is destined to weigh up computational feasibility and spatial scope of context information. Most deep networks \cite{ronneberger2015u,zhou2019unet++,isensee2021nnu,chen2017deeplab}, regardless of their architectures, could effortlessly provide spatial features at different scales and receptive fields. However, it is by no means a trivial task to use these feature layers as part of CRF. Since the networks are trained end-to-end with (partial) segmentation supervision, its internal features were already tailored to the learning target. Therefore, mindlessly feeding feature maps to CRF kernels could intensify incorrect information from the target if the network is not properly rectified. Alternatively, using a separate network to learn representative features for CRF seems plausible, as decoupling from the main backbone creates diversity \cite{luo2022scribble} while avoids strengthening errors. However, this spells another uncertainty in how to design this auxiliary network, and how much resource overhead is expected. 

Motivated by these challenges, we present a straightforward yet effective learning framework. Inspired by recent advances of transformer based segmentation \cite{strudel2021segmenter,hatamizadeh2022unetr}, we create a pluggable spatial self-attention module which could be attached on top of any internal feature layers of a fully convolutional network (FCN). Comparing to end-to-end transformer architecture, this design infuses global contextual information to the main backbone while keeping the efficiency of convolutions. Descended from this module, we construct a similarity metric based on normalized and symmetrized attention. This attentive similarity, together with a distance mapping mask that encodes spatial locations, further participate in a novel regularization loss that imposes consistency between segmentation prediction and visual feature affinity. Comparing to CRF, this loss is efficient since the kernels were pre-computed within the attention module. This attentive similarity objective drives a joint alignment of FCN encoders, attention mapping and model prediction. Ultimately, the entire end-to-end learning framework can be established with the proposed FCN+Attention architecture and a combination of three objectives: partial segmentation loss, a customized masked CRF loss and the attentive similarity loss. The framework is trained in a single stage with scribble annotations, and no pseudo labels or pre-trained feature detectors will be needed. 

The contributions of this work are three-fold. (1) We proposed a pluggable spatial self-attention module on top of FCN for scribble supervised segmentation. The module can work with any backbone architectures and could model interactions of all spatial locations. Stemming from this module, we designed a attentive similarity loss that supports one-stage end-to-end learning. Combining global attention with FCNs is not a entirely new idea \cite{Wang2017NonlocalNN,Li2019GlobalAT,Huang2018CCNetCA}, but to our best knowledge, this is the first work to cultivate a particular design for scribble supervision task, where the novel attentive similarity loss plays a central role in regularization. (2) We customized a masked CRF regularization that drives decent segmentation accuracy and facilitates the capacity of attentive similarity learning. (3) Extensive experiments on public datasets (ACDC and CHAOS) showed that our framework not just out-performs existing state-of-the-art, but also delivers close performance to fully-supervised benchmark. 

\begin{figure}
	\includegraphics[width=\textwidth]{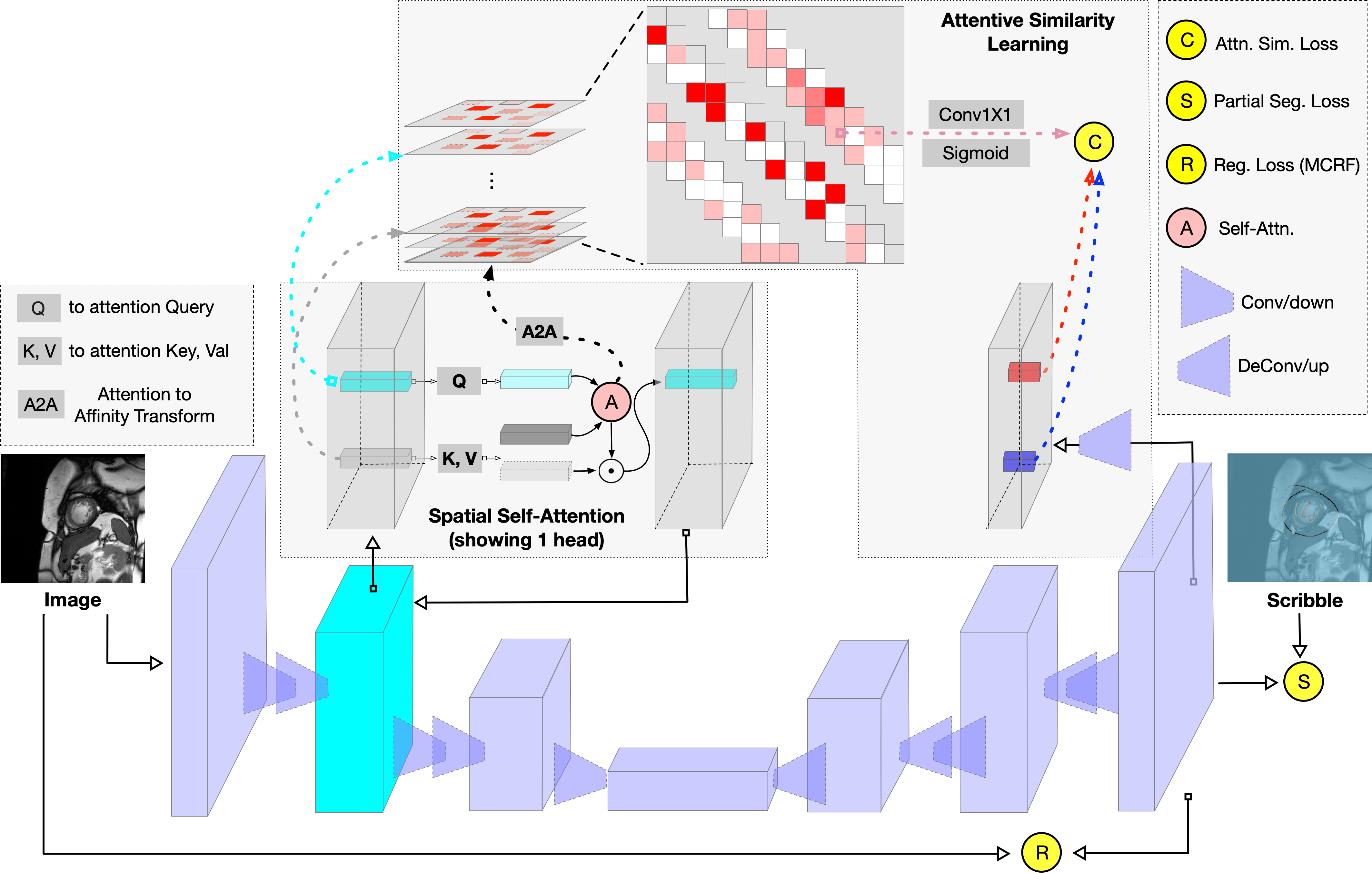}
	\caption{The proposed framework.} \label{fig:model}
\end{figure}

\section{Method}

The proposed method is depicted in Fig. \ref{fig:model}. The backbone network (using a U-shaped one for illustration) is jointly trained with three objectives: the partial segmentation loss, masked CRF loss and the affinity similarity loss. Different to those pseudo label based approaches \cite{lee2013pseudo,grady2006random,lee2020scribble2label,zhang2022cyclemix,luo2022scribble,chen2022scribble2d5}, only labeled pixels along scribbles contributed to the forward and backward operations flows of the (partial) segmentation loss. In addition, we customize a masked CRF for regularization similar to \cite{obukhov2019gated}, which proved effective for suppressing false positives. At the same time, we design an attention guided similarity loss to encourage the consistency between model predictions and learned visual affinity from encoded features. A selected encoder layer will firstly be sent through a spatial self-attention module, then the learned attention will be transformed to construct a similarity mask that signals spatial dependencies. Such a mask will then be combined with a distance map which encodes spatial locations, and eventually formulates the attentive similarity loss. In addition to learning objectives, the spatial self-attention also takes part in architectural improvements by introducing global interactions to CNN features.

\begin{comment}
	The proposed framework consists of a backbone segmentation network (generic encoder-decoder FCN such as U-Net \cite{ronneberger2015u}), a spatial self-attention (SSA) module, a attentive similarity learning (ASL) module and masked CRF (MCRF). SSA can be attached to any internal feature layers, and ASL stems from SSA with attention-to-affinity (A2A) transform. Eventually, the model is jointly training with partial segmentation loss, attentive similarity loss and masked CRF loss.
\end{comment}

\subsection{Scribble Supervision}\label{sec:method:scribble}
The scribble annotation mask consists of a few partially connected pixels that have known class labels $\mathcal{C} = \{0, 1, ..., C\}$, with other pixels untagged.  Let $\Omega_L, \Omega_U$ be the set of labeled and unlabeled pixels, then the label mask is $Y_s[p] = Y[p][p \in \Omega_L] + l_U[p \in \Omega_U]$ where $Y[p]$ is the ground truth label at pixel $p$ and $l_U \notin \mathcal{C}$ is a marker for ``unknown''. Pseudo label based methods resorted to extend labels from $\Omega_L$ to $\Omega_U$. However, the incorrectly propagated labels in $\Omega_U$ will consistently steer the model towards inaccurate predictions. Therefore, we directly use $Y_s$ as the sole supervision. One of such formulations is the partial cross entropy (pCE) \cite{lin2016scribblesup} loss:
\begin{equation}
	l_{seg}(P, Y_s) = -\frac{1}{|\Omega_L|}\sum_{p \in \Omega_L}\log (P_p^{Y_s[p]})
	\label{eq:pce}
\end{equation}
where $P_p^c$ is the model predicted probability that pixel $p$ belongs to class $c$.

\subsection{Masked CRF}\label{sec:method:mcrf}
With partial supervision, the network indeed ``propagates'' labels onto $\Omega_U$ based on learned semantic similarities. However, the prediction could be highly inaccurate around object boundaries due to the inherent deficiency of scribble labels. A most recent work \cite{Zhai2022PASegLF} proposed to use label six border points that effectively alleviated boundary ambiguity. However, scribble annotations normally locate around the target skeleton, which could by no means give such luxurious clues without re-labeling. Conditional random field (CRF) \cite{obukhov2019gated}, a common practice to adjust predicted labels according to elementary spatial features, proved effective for boundary refinement. Specifically, given a similarity metric $s(p, q)$ for any pairs of pixels $\{p \ne q \}$. If $Y[p] \ne Y[q]$, we'd expect $p$ and $q$ to be less similar than otherwise. On the other hand, if $s(p, q)$ has a higher value, then $Y[p]$ and $Y[q]$ will be less likely to differ. This intuition drives the formulation of CRF optimization:
\begin{equation}
	Y^{\star}_{\text{crf}} = \underset{Y}{\text{argmin}} \sum_{p\ne q}s(p, q)\sum_{i \ne j \in \mathcal{C}} [Y[p] = i, Y[q] = j]
	\label{eq:crf}
\end{equation}
Given initial model predictions, solving for $Y$ is an effective post-processing step \cite{chen2017deeplab}. A differentiable loss could be further derived through relaxation:
\begin{equation}
	l_{\text{crf}}(P, I) = \frac{1}{|\Omega|} \sum_{p \ne q \in \Omega}\sum_{i \ne j \in \mathcal{C}} s(p, q)P_{p}^{i}P_{q}^{j}
	\label{eq:crf_loss}
\end{equation}
where $I$ is the image input and P is model prediction. For high resolution medical images, incorporating all pairs of pixels can be quite expensive. Imposing a limited neighboring window of a fixed radius $r$ is thus a common practice. In addition, unwanted pixels can further be excluded to avoid introducing noise. The Gated CRF approach \cite{obukhov2019gated} applied pre-defined masks to zero out the influence from image augmentation induced artifacts. Different to \cite{obukhov2019gated}, we include extrapolated margins from random rotation augmentation into the loss. Such a customized mask helps to mitigate false positives near image borders, while excluding already annotated pixels from scribbles. The masked CRF loss is then formulated as equation \ref{eq:mcrf_loss}:
\begin{equation}
	l_{\text{mcrf}}(P, I) = \frac{1}{|\Omega_U|} \sum_{p\in \Omega_U}\sum_{q \in \Omega_U: 0 < \|q-p\|_{\infty} < r}s(p, q)\sum_{i \ne j \in \mathcal{C}} P_{p}^{i}P_{q}^{j} 
	\label{eq:mcrf_loss}
\end{equation}
Normally, $s(p, q)$ consists of a weighted combination of multiple Gaussian kernels:
\begin{equation}
	s(p, q) = \sum_{k=1}^{K} w^k e^{-\frac{1}{2(\sigma^k)^2}\|f_p - f_q\|_2^2}
	\label{eq:gauss}
\end{equation}
We found that using just a single kernel that encodes intensity and location similarity $f_p = [I_p, x_p, y_p]^{\text{T}}$ could already improve the segmentation accuracy to a reasonable level. 

\subsection{Spatial Self-Attention}\label{sec:method:ssa}
One potential reasons for CRF's success is that it uses the most shallow and localized features: intensity along with spatial location as a complement to deep semantic representations learned within the network. Also, the energy formulation essentially relaxes learning objective from pixel-wise classification to pairwise consistency. A natural attempt then is to use learnable features to replace $f_p$ in equation \ref{eq:gauss}. One possible approach is to introduce an auxiliary network or branch inspired by the elegant design proposed by [dual head wang guotai], the auxiliary architecture could be used to provide richer features for CRF loss. A more challenging approach, as this work focuses on, is to directly exploit the main network without introducing auxiliary overhead. The challenge stems from a paradox: since the network is end-to-end trained with segmentation task, the spatial similarity of internal features should already correspond to those in model prediction. If $P$ is not accurate enough, re-imposing pairwise (dis-)similarity according to $P$ could just intensify both correctly and incorrectly predicted pixels. If otherwise, since learning pixel-wise label is more difficult than similarity, there is no need for introducing the redundancy. Therefore, as opposed to using internal feature maps directly, a new design is needed. 

As mentioned in Sec. \ref{sec:intro}, transformer is better at capturing long range dependencies than CNNs due to self-attention, which could be viewed as embedded affinities among pairs of spatial locations. Instead of switching to transformer based backbone, we tailor self-attention into a plugable module that could be attached to any feature layers of arbitrary CNN architectures. Let $f\in \mathbb{R}^{hw\times d}$ be the flattened feature map with $h, w$  be the spatial resolution and $d$ the feature dimension, then the $i^{\text{th}}$ attention head is constructed by three individual linear transforms, producing query $Q_i=\text{MLP}_{d\to d_q}(f)\in \mathbb{R}^{hw\times d_q}$, key $K_i=\text{MLP}_{d\to d_k=d_q}\in \mathbb{R}^{hw \times d_k}$ and value $V_i =\text{MLP}_{d\to d_v} \in \mathbb{R}^{hw \times d_v}$. The attention then will be  $A_i= Q_iK_i^\text{T} / \sqrt{d_k} \in \mathbb{R}^{hw\times hw}$ and the attended feature will be $a_i = \text{softmax}(A_i)V_i \in \mathbb{R}^{hw\times d_v}$. The concatenated multi-head features $a=\text{concat}(a_i) \in \mathbb{R}^{hw \times d_vn)}$ will then go through layernorm and MLP with skip connections: $a = a + \text{MLP}_{d_v\to d_v}(\text{LayerNorm}(a))$. If there are multiple identical layers of these operations in a sequence, the output feature of current block will be the input of the next. 

In our context, the attended feature will eventually need to be compressed back to the original dimension with $a = \text{MLP}_{ndv \to d}(a)$ for follow-up convolutions. Such seemingly straightforward module brings two benefits: first, it encapsulates global long-term dependencies within the CNN backbone, and thus increased the scope of receptive field to the entire spatial domain. Second, the learned attention $A_i$ bears clues for pair-wise similarity among different locations. If we could take advantage $A_i$ into a regularization loss such as equation \ref{eq:mcrf_loss}, then $s(p, q)$ can be obtained without heavy $O(h^2w^2)$ computations. 

\subsection{Attentive Similarity Learning}\label{sec:method:acl}
This section will describe how to transform the attention into a regularization loss and why our design could work in a way free from the earlier mentioned paradox. 

Similar to attended features, we need to concatenate $A_i$ from all $n \geq 1$ heads and all $L \geq 1$ layers: $A = \text{concat}(A_i)\in \mathbb{R}^{hw \times hw \times nL}$. Since $Q$ and $K$ are different transforms, $A$ is normally asymmetric. Next, we use a soft-max to normalize rows in $A$ for the reasons to be explained later: $A = \text{softmax}(A)$. Then, similar to [aff loss], we derive a symmetric version through: $\bar{A} = A + A^{T} \in \mathbb{R}^{hw \times hw \times nL}$ where the transpose is taking over first two dimensions. Eventually, a $1\times 1$ convolution is used to compress the channel dimension and transform $\bar{A}$ into pairwise similarity metric $S = \text{sigmoid}(\text{Conv}^{1\times 1}_{nL \to 1}(\bar{A})) \in \mathbb{R}^{hw \times hw}$. 

It might seem convenient to use $s(p, q) = S[p, q]$ directly in equation \ref{eq:mcrf_loss} to replace the hand-crafted kernel. However, doing so will deteriorates model performance since $S$ does not encode effective positional information. Though convolution operation with zero-padding could leak partial location signals implicitly [segformer, segformer-69], it is not sufficient in our context that requires accurate distance mapping. Without $x, y$ location included, there will be a risk of introducing extraneous false-positives at remove locations with large similarity values in $S$. We encode location information explicitly with a distance decay mapping: $\mathcal{M} \in \mathbb{R}^{hw \times hw}$  with $\mathcal{M}[p, q] = \exp\{-\|p_{x, y} - q_{x, y}\|^2_2 / 2\sigma^2\}$. In addition, the network layer attached to the attention module normally have lower resolution than predicted mask $P$, in this case, we down-sample $P$ to $h\times w$: $\tilde{P} = \text{Interpolate}(P, h \times w)$. To this end, we have the regularization loss from attention module in equation \ref{eq:atn_loss}:

\begin{equation}
	l_{\text{atn}}(P, S) = \frac{1}{|\Omega_U|} \sum_{p\in \Omega_U}\sum_{q \in \Omega_U: 0 < \|q-p\|_{\infty} < r}\mathcal{M}[p, q]S[p, q]\sum_{i \ne j \in \mathcal{C}} \tilde{P}_{p}^{i}\tilde{P}_{q}^{j} 
	\label{eq:atn_loss}
\end{equation}

Though taking a similar form, this loss is fundamentally different to Masked CRF one in equation \ref{eq:mcrf_loss}. This is a joint optimization of $P$ and $S$; during back-propagation, both the main network and the attention branch will take gradient updates. An improved $P$ will drive the production of better $S$, which will eventually refine the representation of in earlier features and attention. Other the hand, when sitting on earlier network layers, $S$ infuses shallow visual information into the objective, which echoes the effect of vanilla CRF. In addition, the learnable parameters in attention block promote information diversity that prevents the downward spiral from intensifying incorrect predictions. In addition, this loss is implicitly contrastive due to the critical soft-max operation on learn attention $A$. Suppressing similarities among pairs with distant $P$ will automatically encourage higher similarity among pairs with close $P$. Otherwise, a naive optimization would be to set all $S[p, q]$ to zero. In summary, this attentive similarity loss solved the aforementioned challenge in designing CRF-like regularization within a single network. We will show demonstrate effectiveness in experiment. 

Ultimately, the network will be trained by optimizing the following objective: \cite{bernard2018deep}

\begin{equation}
	l = l_{\text{seg}}(P, Y^s) + \lambda_{\text{mcrf}}l_{\text{mcrf}}(P, I) + \lambda_{\text{atn}}l_{\text{atn}}(P, S)
	\label{eq:total_loss}
\end{equation}

\section{Experiment and Results}
\subsection{Datasets}
We evaluate the proposed method and existing state-of-the-arts methods on two datasets, ACDC \cite{bernard2018deep,luo2022scribble} and CHAOS \cite{kavur2021chaos,valvano2021learning}. The ACDC datasets includes 200 3D cine-MRI images obtained from 100 patients (two images from each patient). Ground truth annotations are provided for end-diastolic (ED) and end-systolic (ES) cardiac phases for three foreground targets: right ventricle (RV), myocardium (Myo), and left ventricle (LV). Scribble annotations were directly adopted from a recent release in \cite{valvano2021learning}. The CHAOS dataset has 20 3D abdominal MR images of 20 subjects (one image per subject), full segmentation masks are provided for liver, kidneys and spleen. We choose the T1 in-phase in this paper. Since no scribbles were provided, we simulated scribble annotations for foreground and background classes similar to the steps in \cite{valvano2021learning}. 

To acquire the scribble annotation in CHAOS T1 Phase, we followed the methods outlined in the research article by Valvano et al.\cite{valvano2021learning}. The scribble annotations for each organ were obtained through standard skeletonisation and interactive erosion techniques, whereby the border pixels were continuously eroded until the point of loss of connectivity. Additionally, to create a boundary that closely approximates the periphery of aimed object and to reduce the cost of model learning error correction, the skeleton of the outside boundary was constrained. Specifically, we generated the largest convex hull that could encapsulate all organs and then expanded the convex hull by 5 pixels to serve as an additional boundary for the overlapping background area. Pixels falling outside of these defined boundaries were left unclassified as belonging to either organ or background.

The generated scribble data will be released upon publication.

For both dataset, we split it into training and evaluation sets by patient. For ACDC, 20 patients (40 images) were allocated for evaluation and the rest 80 patients (160 images) for training. For CHAOS, 7 subjects/images were used for evaluation and the rest 14 subjects/images for training. All methods will be trained and tested on the identical set-up. For reproducibility, the patient ids corresponding to this partition will be given upon code release. 

\subsection{Implementation}
We choose 2D U-Net \cite{ronneberger2015u} as opposed to 3D FCNs as the backbone network. Since that on both dataset the thickness between slices is $\sim$6 times larger than pixel spacing, also that scribbles were delineated on 2D slices instead of 3D volumes \cite{luo2022scribble,valvano2021learning}. The proposed self-attention module were grafted onto the third convolutional layer of encoder. The detailed network configurations will be included in supplementary materials. All experiments were run with PyTorch 1.7.1 \cite{Paszke2019PyTorchAI} environment on 8 RTX A6000 GPUs. During training, we standardize image intensity to the range of $[0, 1]$ for each 3D volume. Then, each slice goes through augmentation steps such as random rotation/flipping following a resizing to $256\times 256$. This process is similar to \cite{luo2022scribble}, but we took more stringent augmentations on CHAOS since flipping and $90^{\circ}$ rotation could confuse left/right kidneys. The network is trained end-to-end with stochastic gradient descent and polynomial learning rate decay schedule. We set both $\lambda_{\text{atn}}$ and $\lambda_{\text{mcrf}}$ to 0.1 and sensitivity analysis will be provided in supplementary material. During testing, 2D predictions were stacked to produce 3D masks. No post-refinement steps were adopted. The 3D Dice Coefficient and 95\% (DSC) Hausdorff Distance (HD95) are used for evaluation. Note that HD95(mm) should be computed with pixel spacing and thickness considered. 

\subsection{Results}\label{exp:res}
\subsubsection{Comparing with Other Methods}
\begin{table}[t]	
	\caption{Eval. on ACDC. Displayed \texttt{mean(std)} of DSC and HD95(mm) across all test samples, for three different foreground classes (RV, Myo, LV) respectively as well as avg. across all classes. Top 2 best metrics (DSC$\uparrow$, HD95$\downarrow$) including ties will be shown in \textbf{bold}, top metric without ties will be shown in \textbf{\textit{bold-italic}}}  \label{tab:acdc}
	%	\begin{tabular}{l|l|l|l|l|l|l|l|l}
		%    \begin{tabular}{lccccccccc}
			\resizebox{\textwidth}{!}{
				\begin{tabular}{lllllllll}
					\hline
					\multirow{2}*{Method} &  \multicolumn{2}{c}{\textbf{RV}}  & 	\multicolumn{2}{c}{\textbf{Myo}} & \multicolumn{2}{c}{\textbf{LV}} & \multicolumn{2}{c}{\textbf{Avg.}} \\
					
					%		  & \makecell[c]{$DSC$} & \makecell[c]{$HD_{95}$} & \makecell[c]{$DSC$} & \makecell[c]{$HD_{95}$} & \makecell[c]{$DSC$} & \makecell[c]{$HD_{95}$} & \makecell[c]{$DSC$} & \makecell[c]{$HD_{95}$} \\
					& DSC & HD95 & DSC & HD95 & DSC & HD95 & DSC & HD95 \\
					\hline
					Fully & 0.89(0.079) & 10.47(20.17) & 0.88 (0.028) & 2.21(1.91) & 0.96 (0.009) & 2.21(2.07) & 0.91 (0.031) & 4.96(6.97) \\
					\hline
					\cline{1-9}
					pCE\cite{lin2016scribblesup} & 0.58(0.16) & 186.25(28.7) & 0.48(0.12) & 166.67(17.67) & 0.79(0.10) & 167.41(18.24) & 0.62(0.10) & 173.44(18.56)\\
					RW \cite{grady2006random} & 0.82(0.11)& 12.7(24.72) & 0.7(0.049) & 8.01(2.43) & 0.89(0.036) & 10.4(14.04) & 0.8 (0.043) & 10.37(12.97 )  \\
					USTM \cite{liu2022weakly} & 0.75(0.13) & 124.93(64.88) & 0.69(0.079) & 141.66(24.83) & 0.82(0.068) & 153.36(28.2) & 0.75(0.069)&139.98(30.25)  \\
					
					S2L \cite{lee2020scribble2label}&  0.84(0.095) & 17.61(31.27) & 0.8(0.056) &46.52(53.71) &0.92(0.039) & 53.31(58.32) & 0.86(0.041) & 39.14(37.07) \\
					MS \cite{kim2019mumford} & 0.86(0.075) & \textbf{11.31(18.85)} &	\textbf{0.83(0.043)} & 14.14(34.46) & \textbf{0.94(0.027)} & 17.42(38.39) & \textbf{0.88(0.033)} 	& 14.29(25.69) \\
					EM \cite{tang2018normalized} &  0.83(0.83)	& 24.07(42.55)	& 0.81(0.051)	& 44.18(51.55) &	0.92(0.035) & 36.19(48.14)	& 0.85(0.043) & 34.81(35.71) \\
					GCRF \cite{obukhov2019gated} & \textbf{0.87(0.083)}	& 11.7(21.51) &	\textbf{0.83(0.036)} & 15.76(36.28)	& \textbf{0.94(0.018)}	& 13.64(35.06) & \textbf{0.88(0.034)} & 13.70(24.77)  \\	
					Contour \cite{jurdi2021surprisingly} & 0.61(0.12) & 192.85(27.55) & 0.56 (0.084) & 162.88(16.81)	& 0.79(0.066) & 172.08(20.96) & 0.65(0.071) & 175.94 (18.50)  \\	
					DBr \cite{luo2022scribble} & 0.86 (0.085) & 15.98(30.52) & \textbf{0.83(0.041)}	& 20.72(40.81) & 0.93	(0.023)	& 24.51(43.77)	& 0.87(0.034) & 20.40(30.45) \\	
					\hline
					Ours w/o ASL & \textbf{0.87(0.097)} &13.79(29.16) &\textbf{0.83(0.042)} &\textbf{\textit{3.45(1.99)}}	&\textbf{0.94(0.027)} &\textbf{\textit{4.16(6.38)}} &\textbf{0.88(0.040)} &\textbf{\textit{7.13(10.02)}}\\
					Ours & \textbf{0.87(0.079)}	&\textbf{\textit{10.72(19.34)}}	&\textbf{0.83(0.038)}	&\textbf{\textit{3.29(1.84)}}	&\textbf{0.94(0.016)}	&\textbf{\textit{5.13(8.86)}}	&\textbf{0.88(0.032)} &\textbf{\textit{6.38(8.53)}}\\											
					\hline
				\end{tabular}
			}
		\end{table}

		% CHAOS
		\begin{table}[t]
			\caption{Eval. on CHAOS. Displayed \texttt{mean(std)} of DSC and HD95(mm) across all test samples, for four different foreground classes (LIV, RK, LK, SPL) respectively as well as avg. across all classes. Top 2 best metrics (DSC$\uparrow$, HD95$\downarrow$) including ties will be shown in \textbf{bold}, top metric without ties will be shown in \textbf{\textit{bold-italic}}}\label{tab:chaos}
			\resizebox{\textwidth}{!}{
				\begin{tabular}{lllllllllll}
					\hline
					\multirow{2}*{Method} &  \multicolumn{2}{c}{\textbf{LIV}}  & \multicolumn{2}{c}{\textbf{RK}} & 
					\multicolumn{2}{c}{\textbf{LK}} & \multicolumn{2}{c}{\textbf{SPL}} & \multicolumn{2}{c}{\textbf{Avg.}}\\
					
					& DSC & HD95 & DSC & HD95 & DSC & HD95 & DSC & HD95 & DSC & HD95\\
					\hline
					
					Fully & 0.92(0.016)	&10.0(3.88) &0.9(0.030) &20.72(16.89)	&0.9(0.061) &13.2(14.12) &0.83(0.090) &12.2(8.35) &0.89(0.041)	&14.03(6.20) \\
					\hline
					\cline{1-11}
					pCE\cite{lin2016scribblesup} & 0.85(0.025) &	12.24(2.13)	& 0.71(0.095) &	51.28(39.81) &	0.72(0.110) &	46.48(31.35) &	0.71(0.1)	& 39.19(14.22)&	0.75(0.076)	&37.3(11.90) \\
					RW \cite{grady2006random} & 0.762(0.028) &22.74(15.54) &0.634(0.101) &44.57(19.93) &0.676(0.152) &18.05(9.95) &0.725(0.120) &30.94(19.47) &0.699(0.090) &29.07(6.41) \\
					USTM \cite{liu2022weakly}& 0.81(0.024)	& 13.78(0.69)	& 0.7(0.088)	&30.29(12.45)	& 0.72(0.12) & 45.76(34.27)	& 0.69(0.19)	&33.32(20.28)	& 0.73(0.1) &30.79(13.49) \\					
					S2L \cite{lee2020scribble2label}&  0.82(0.024)	&34.73(16.65)	&0.7(0.065)	&72.74(56.45)	&0.67(0.14) &92.38(67.07)	&0.64(0.21)	&84.91(28.33)	&0.71(0.09) &71.19(15.45) \\					
					MS \cite{kim2019mumford} &0.64(0.035) &25.2(1.42) &0.65(0.068) &54.13(18.50)	&0.63(0.1) &34.96(31.64)	&0.62(0.21) &25.59(20.13) &0.64(0.089) &34.97(14.22) \\
					EM \cite{tang2018normalized}&0.84(0.022) &11.51(0.88) &0.69(0.084) &28.35(14.39) &0.71(0.12) &39.31(43.56) &0.71(0.14) &28.38(11.69) &0.74(0.083)	&26.89(11.28) \\
					GCRF \cite{obukhov2019gated}& \textbf{0.92(0.022)} &\textbf{8.97(2.61)} &\textbf{0.87(0.043)} &25.02(15.63) &\textbf{0.89(0.053)} &10.79(9.19) &\textbf{0.85(0.062)} &\textbf{11.13(7.37)} &\textbf{0.88(0.037)}	&\textbf{13.98(4.44)} \\	
					Contour \cite{jurdi2021surprisingly}& 0.84(0.022) &13.29(2.12) &0.66(0.1) &23.84(13.83) &0.71(0.15) &14.85(8.13) &0.72(0.12) &28.72(21.97) &0.73(0.096)	&20.18(7.01)  \\	
					DBr \cite{luo2022scribble}& 0.89(0.020) &11.19(4.09) &0.85(0.028) &32.33(25.68) &0.84(0.082) &15.13(10.56) &0.79(0.12) &12.77(7.65) &0.84(0.054)	&17.85(6.99) \\
					\hline
					Ours w/o ASL & \textbf{0.92(0.018)} &14.46(8.65)	&0.86(0.030) &\textbf{\textit{19.79(14.66)}}	& \textbf{0.89(0.038)} &\textbf{\textit{8.55(6.06)}} &0.82(0.1) &13.18(7.91) &\textbf{0.87(0.041)} & \textbf{13.99(5.51)} \\
					Ours &  \textbf{\textit{0.93(0.017)}} &\textbf{\textit{8.19(1.56)}} &\textbf{0.87(0.030)} & \textbf{\textit{18.55(17.57)}} &\textbf{0.89(0.052)} &\textbf{\textit{7.79(5.23)}} &\textbf{0.84(0.089)} &\textbf{\textit{10.21(8.18)}}	&\textbf{0.88(0.040)} &\textbf{\textit{11.18(5.00)}}\\

					\hline
				\end{tabular}
			}
		\end{table}
		
		We compare our method with 8 existing methods \cite{grady2006random,liu2022weakly,lee2020scribble2label,kim2019mumford,Grandvalet2004SemisupervisedLB,obukhov2019gated,jurdi2021surprisingly,luo2022scribble} for scribble supervised segmentation and the supposed lower/upper benchmarks. In particular, training the network with partial cross entropy (pCE) \cite{lin2016scribblesup} loss alone can be treated as a baseline benchmark. At the other end of the spectrum, training with fully annotated masks (Fully) should provide upper bound performance since no regularization/pseudo label could produce a stronger supervision than pixel-level ground truth. 
		
		Among the 8 methods, RW (random walker) \cite{grady2006random}, USTM \cite{liu2022weakly}, S2L (Scribble2Label) \cite{lee2020scribble2label} and the most recent DBr (dual-branch) \cite{luo2022scribble} leveraged pseudo labels in one way or another. In particular, both USTM \cite{liu2022weakly} and DBr \cite{luo2022scribble} adopted double network/branches to promote robustness. Besides, the rest of four methods, MS (Mumford-Shah loss) \cite{kim2019mumford}, EM (Entropy Minimization) \cite{Grandvalet2004SemisupervisedLB}, GCRF (GatedCRFLoss) \cite{obukhov2019gated} and Contour (contour loss) \cite{jurdi2021surprisingly} explored regularization terms with pCE training. Another recent work, Scribble2D5 \cite{chen2022scribble2d5} also showed exceptional performances, however, it is not included in our comparison since they rely on external dataset to pre-train an edge detector, which may induce uncertainty in both fairness and interpretability of the experiment. 
		
		Tables \ref{tab:acdc} and \ref{tab:chaos} show the results on ACDC and CHAOS respectively. For both datasets, our framework (last row in the tables) delivers the best DSC and HD95. Some methods such as Dual Branch \cite{luo2022scribble}, Mumford-Shah loss \cite{kim2019mumford} and Gated CRF \cite{obukhov2019gated} give the same level of DSC on particular organs, but our HD95 is significantly lower than all others. For CHAOS, our method can even produce lower HD95 and same level of DSC (on LIV, LK, SPL and average) comparing to the fully supervision benchmark. This indicates that with masked CRF and attentive similarity learning, our method is better at preserving target boundaries and reducing false positives. This conclusion can be further verified in Sec. \ref{sec:res:qualitative}. For medical images, precisely predicting object boundaries is critical in extracting subtle 3D structures with clinical significance. We also observed that in both tables, Gated CRF \cite{obukhov2019gated} and our method yield equal DSC. This may be caused by two reasons that also reveal the inherent mechanisms: (1) limited by sparse annotation, there is an upper limit of DSC below fully supervised benchmark; the tables (especially table \ref{tab:chaos}) show that both methods approach this upper limit, and therefore the similar DSC. (2) the role of masked CRF (Sec. \ref{sec:method:mcrf}) which resembles Gated CRF \cite{obukhov2019gated} should be to push the prediction accuracy to a reasonable level, upon which attentive similarity learning (Sec. \ref{sec:method:acl}) takes over to optimize the boundaries. The ASL module is fundamentally different to masked CRF, and cannot be used to replace CRF. This verified our statements in Sec. \ref{sec:method:acl}.  
		
		We noticed that in table \ref{tab:acdc}, both fully supervised learning and our method gives relatively higher HD95 when predicting RV comparing to other targets. However, this is not the case for other methods. We will investigate further whether there's a more consistent resemblance between our method and fully supervised learning in supplementary material through cross validation. 
		
		\subsubsection{Ablation Study}
		It might seem reasonable to investigate individually the importance of three essential modules, SSA (Sec. \ref{sec:method:ssa}), MCRF (Sec. \ref{sec:method:mcrf}) and ASL (\ref{sec:method:acl}) of our framework. The three modules need to work together to guarantee the top performance. MCRF helped to boost the accuracy in target localization and shape (measured by DSC), then SSA and ASL could then show their capacities in reducing false-positives and preserving boundaries. In addition, there's no way to keep ASL without SSA since ASL was directly derived from SSA. 
		
		Therefore, we only examine how the performance could change if we remove ASL. From both tables (second row from bottom), removing ASL would make HD95 worse, for all organs and avg. with one exception for LV (table \ref{tab:acdc}). This further verifies our statement about ASL's role on boundary preservation. 
		
		Moreover, even without ASL, our DSC and HD95 still stay on the best level for most part of ACDC and CHAOS, with three exceptions including ACDC-RV, CHAOS-LIV and CHAOS-SPL. This further indicates that both SSA and ASL contributed to reducing boundary inconsistencies. 
		
		\begin{figure}
			\includegraphics[width=\textwidth]{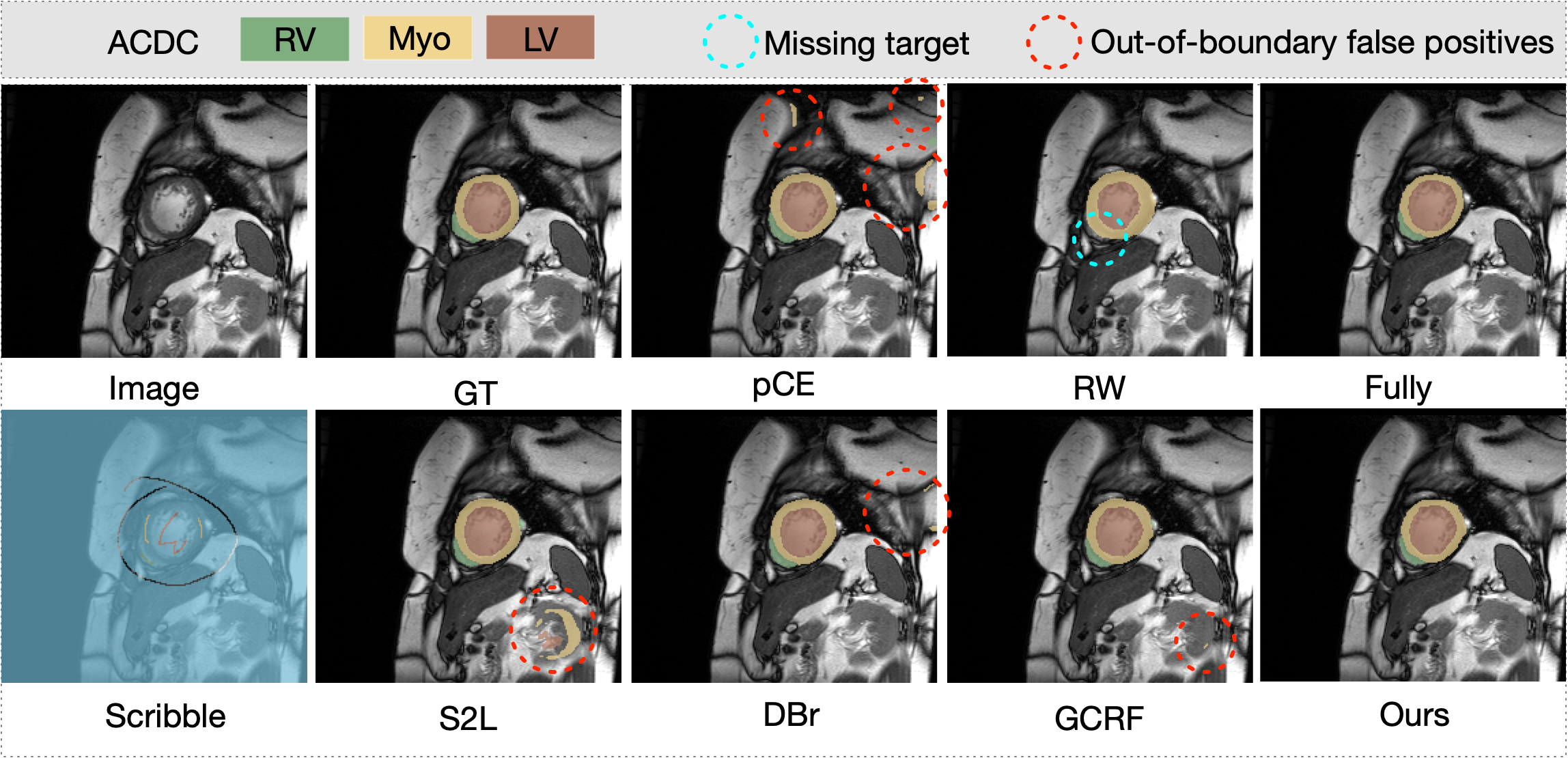}
			\caption{Segmentation predictions on ACDC slice, GT is ground truth mask. We identify with circles significant false-positives and missing areas (false-negatives)} \label{fig:acdc2d}
		\end{figure}

		\begin{figure}
			\includegraphics[width=\textwidth]{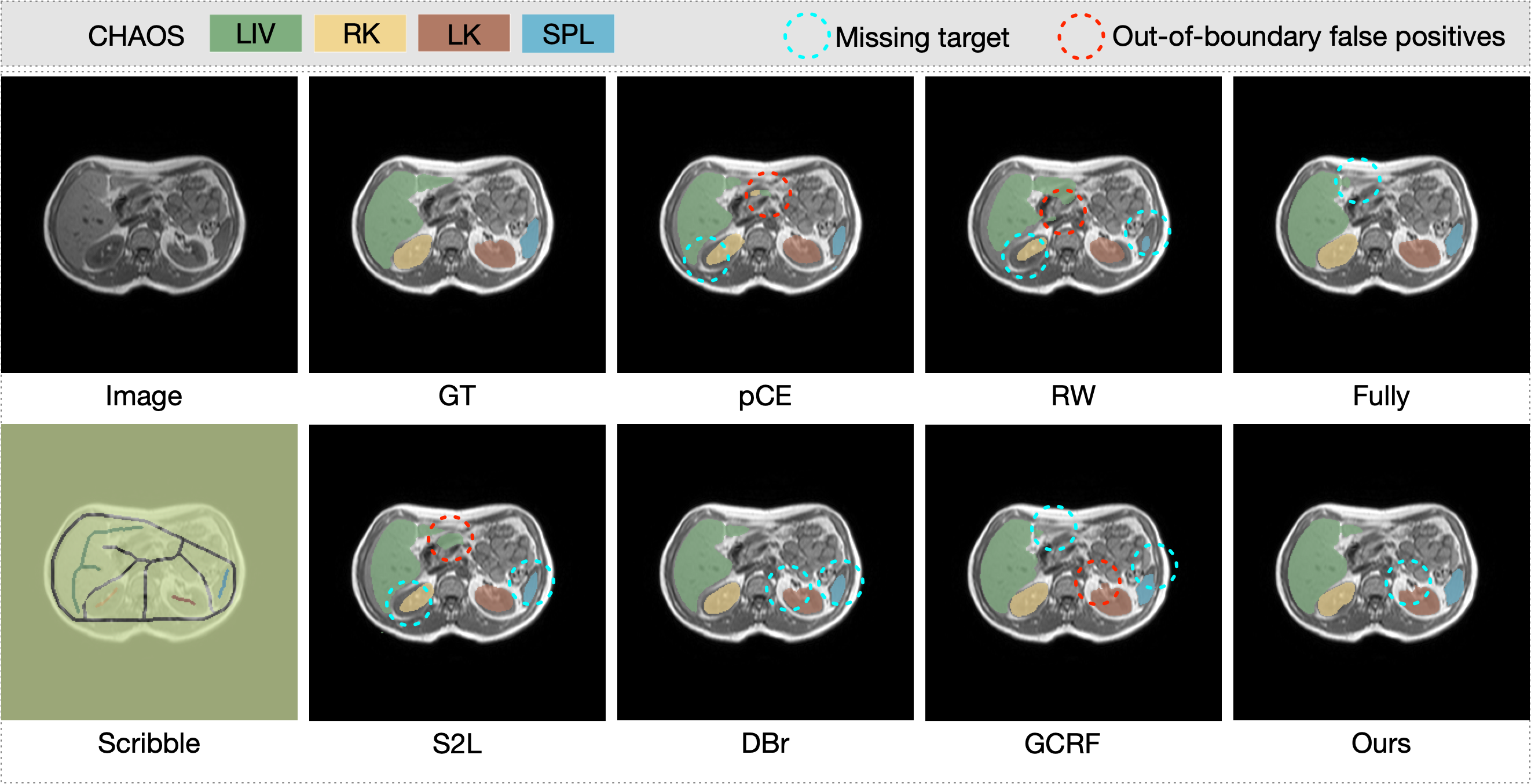}
			\caption{Segmentation predictions on CHAOS slice, GT is ground truth mask. We identify with circles significant false-positives and missing areas (false-negatives)} \label{fig:chaos2d}
		\end{figure}
		
		\subsection{Qualitative Analysis}\label{sec:res:qualitative}
		We visualize segmentation results along with scribbles/ground truth labels from representative 2D slices in Figs. \ref{fig:acdc2d} and \ref{fig:chaos2d}. In Fig. \ref{fig:acdc2d}, our and fully supervised model provide the most similar predictions to the ground truth. In addition, false positives can be effectively mitigated by our approach comparing to other methods. This in turn demonstrates the conclusion observed in tables \ref{tab:acdc} and \ref{tab:chaos}. In Fig. \ref{fig:chaos2d}, we see that aside from suppressing false positives, our method also preserves target boundary better while all other methods give cupped boundaries for some target. Comparing to fully supervised learning, our method missed out a small part of LK, but sufficiently predicted the tipping part of LIV missed out by Fully. 
		
		\section{Discussion}
		
		Motivated by existing challenges in scribble supervised medical image segmentation, we presented a end-to-end learning framework based on spatial self-attention (SSA), masked conditional random field (MCRF) and attentive similarity learning (ASL). Spatial self-attention module can be plugged on any internal layers of FCN, which increased representation power by infusing global interactions. Attentive similarity learning branches from SSA to impose consistency between visual feature affinity and prediction. In addition, MCRF provided decent segmentation accuracy that in turn facilitate the capability of SSA and ASL. Experiments on two common public datasets (ACDC and CHAOS) showed that the propose method outperformed nine popular strategies on overall performance, particularly on the significantly reduced HD95 values. We will extend our study to more weakly supervised segmentation scenarios in the future. 

\paragraph{Acknowledgment}
This work was supported in part by the Key-Area Research andDevelopment Program of Guangdong Province grant 2021B0101420005,the Shenzhen Natural Science Fund (the Stable Support Plan Program20220810144949003), the Key Technology Development Program ofShenzhen grant JSGG20210713091811036, the Shenzhen Key LaboratoryFoundation grant ZDSYS20200811143757022, the SZU Top RankingProject grant 86000000210, and the National Natural Science Foundationof China grant 62301326.

\bibliographystyle{unsrt}  
\bibliography{references}  %%% Remove comment to use the external .bib file (using bibtex).


\end{document}